\documentclass[11pt]{article}

\usepackage[margin=1in]{geometry}
\usepackage[T1]{fontenc}
\usepackage[utf8]{inputenc}
\usepackage{lmodern}
\usepackage{microtype}
\usepackage{graphicx}
\usepackage{booktabs}
\usepackage{placeins}
\usepackage{url}
\usepackage[hidelinks]{hyperref}
\usepackage[authoryear]{natbib}

\title{Voice Under Revision: Large Language Models and the Normalization of Personal Narrative}

\author{
Tom van Nuenen\\
Social Sciences D-Lab, University of California, Berkeley\\
\texttt{tomvannuenen@berkeley.edu}
}

\date{}

\begin{document}

\maketitle

\begin{center}
\small
\textit{Preprint. This manuscript is currently under review.}
\end{center}



\abstract{%
\textbf{Purpose}
This study characterizes the stylistic shifts introduced by LLM rewriting of personal narratives, evaluates their consistency across frontier models, and assesses whether voice-preserving prompts attenuate them.\\[6pt]
\textbf{Design/methodology/approach}
300 personal narratives are submitted to three frontier LLMs (GPT-5.4, Claude Sonnet 4.6, Gemini 3.1 Pro) under three prompt conditions: generic improvement, rewrite-only (no evaluative language), and voice-preserving. Change is measured across 13 markers grounded in computational stylistics.\\[6pt]
\textbf{Findings}
All three models push the 13 selected markers in the same direction: function words and first-person pronouns decrease; vocabulary diversity, word length, and punctuation elaboration increase. This directional convergence holds whether the prompt says ``improve'' or merely ``rewrite.'' Voice-preserving prompts reduce effect magnitude by 32\% but preserve direction. Stylometric analysis shows practical consequences: texts converge in PCA-projected feature space, and the ability to match rewrites back to source texts drops substantially. Narrative stance markers reveal two linked shifts: from embedded to distanced narration, and from explicit causal reasoning to compressed abstraction.\\[6pt]
\textbf{Originality}
Contemporary LLMs exert a consistent directional pull toward a more polished, less situated register, systematically altering the stylometric and voice features on which DH interpretation often depends.\\[6pt]
\textbf{Contribution to the field of Digital Humanities}
These findings position LLM revision as a consequential new form of textual mediation for digital humanities. The features most strongly affected include those central to many computational stylometry methods, raising questions about corpus integrity, authorship attribution, and the interpretive status of AI-mediated texts.}


\maketitle

\section{Introduction}
\label{sec:introduction}

Texts never reach analysis in a pristine state. DH scholarship has long shown that the textual surface available for interpretation is often shaped by transcription, OCR correction, digitization, and editorial normalization rather than preserved as a transparent record of production \citep{kirschenbaum2013textual, hinrichs2019language, trace2022archives}. The question of what happens to text between production and analysis is, in this sense, a longstanding one. 

The current article extends that question to a new site of mediation: AI-assisted writing. Grammar checkers, predictive text, smart compose features, and chat-based writing assistants now intervene routinely in everyday writing. Where earlier tools applied rule-based corrections, current systems operate probabilistically and at the level of meaning, recasting phrasing, compressing or elaborating content, and reshaping textual form. 

This matters especially for personal narratives, where readers are more readily invited to treat textual features as traces of a real, situated self rather than as properties of a fictional construct. The textual surface of such narratives---including features such as contractions, lexical patterning, and first-person referencing---is often read as evidence of voice and lived experience. AI-assisted revision in personal narratives, therefore, exemplifies a methodological problem for DH: it can alter precisely those features scholars may later treat as analytically meaningful.

Prior work suggests that AI-assisted revision can regularize style, but it is not yet well established whether LLM rewriting systematically shifts the specific linguistic features that carry evidential weight in narrative analysis, or whether such shifts converge across models. To examine this, the study submits 300 personal narratives from the EmpathicStories corpus \citep{shen2023empathicstories} to three frontier LLMs (GPT-5.4, Claude Sonnet 4.6, and Gemini 3.1 Pro) under three prompt conditions: generic improvement, rewrite-only, and voice-preserving improvement. Together, these conditions distinguish between broad revision, non-evaluative rewriting, and explicit instruction to preserve authorial voice.\footnote{While literary and narrative theory often treats voice as a complex amalgamation of identity and social location, for the purposes of this empirical study ``voice'' is operationalized as variance across the selected stylometric markers. These markers were not chosen to define voice but are grounded independently in prior stylometric and register research \citep{stamatatos2009survey,biber1988variation}.} The analysis measures change across 13 markers grounded in computational stylistics and register studies, including function word ratios, vocabulary richness measures, and markers of informal register.

The central finding is directional convergence. All three models push all 13 selected markers in the same direction. Function words and first-person pronouns decrease; vocabulary diversity, word length, and punctuation elaboration increase. This pattern holds whether the prompt says ``improve'' or merely ``rewrite,'' suggesting a shared directional tendency in current LLM rewriting. Voice-preserving prompts reduce effect magnitude but largely preserve the direction of transformation. The result is a consistent pull toward a more polished, less situated register, in which contractions become full forms, repetition becomes variation, and informal phrasing becomes standard prose. Some rewrites also exhibit \textit{stylization}, adding expressive markers (dramatic emphasis, literary phrasing) absent from the original. As this stylometric convergence raises the question of what this compression means for narrative itself, the study also examines how rewriting affects narrative stance along dimensions of narrated positioning (embedded vs.\ distanced) and explanatory style (explicit vs.\ compressed). 

This study makes three contributions. First, it shows that user control is asymmetric. Voice-preserving prompts effectively suppress additive changes (stylization, elaboration of individual texts) but are less effective at preserving features the model tends to remove (contractions, first-person density, function words). Second, the study documents cross-model convergence: GPT-5.4, Claude Sonnet 4.6, and Gemini 3.1 Pro---developed by different companies with different training mixtures and alignment techniques---push all 13 selected markers in the same direction, though with substantial variation in magnitude (mean $|d|$ ranges from 0.60 to 1.45 across models). Third, the study demonstrates practical consequences for DH methods: rewritten texts converge in stylometric space, and source-text recoverability drops substantially (from 14.3\% for GPT-5.4 to near-chance levels for Claude and Gemini). At the level of narrative stance, LLM rewriting produces two linked shifts: eventive markers decrease while retrospective framing increases (a shift from embedded to distanced narration), and causal connectives decrease while abstraction increases (a shift from explicit reasoning to compressed interpretation). The features most strongly affected are those central to stylometric analysis; the stance dimensions most affected are those through which narrators position themselves toward their own experience. Together, these findings suggest that normalization is a structural feature of current LLM rewriting.

\section{Related Work}
\label{sec:related_work}

This article sits at the intersection of digital humanities work on textual mediation, research on stylistic evidence in narrative analysis, and emerging scholarship on AI-assisted writing. While scholars have shown that texts are shaped by editorial, archival, and computational interventions, less is known about whether LLM-based revision systematically alters the linguistic features that humanists later treat as evidence of voice, perspective, and narrative form.

\subsection{Textual Mediation and Corpus integrity}
Texts are shaped by the processes through which they are collected, transcribed, digitized, and cleaned. Born-digital textuality alters the archival conditions under which literary evidence is encountered \citep{kirschenbaum2013textual}; preprocessing decisions can materially affect downstream analysis \citep{kunilovskaya2021text}; archival processing itself functions as a form of infrastructural maintenance that rarely registers as intervention \citep{trace2022archives}. Recent work on digitization and corpus construction has further shown how the composition and representativeness of collections shape the claims that can be made from them \citep{zaagsma2023digital,makela2026opening}. The present study extends this line of inquiry to AI-assisted rewriting, asking whether LLM revision constitutes a kind of textual mediation that operates at the level of corpus integrity and evidentiary status.

\subsection{Voice, Style, and Narrative Markers}
The stakes of such mediation are methodological as well as archival, because linguistic patterning serves as evidence across multiple DH traditions. Computational stylistics operationalizes style through vocabulary distribution, syntactic structure, and punctuation usage \citep{argamon2010computational,herrmann2021computational}; authorship attribution treats such features as relatively stable individual signatures \citep{stamatatos2009survey,eder2016stylometry,mikros2024crosslinguistic}; work on personal narrative links linguistic form to self-presentation, memory, and narrative organization \citep{pennebaker1999linguistic,pennebaker2003psychological,labov1967narrative}. The underlying claim is not that any single feature transparently reveals voice, but that patterning is analytically meaningful. This position is reinforced by recent arguments that stylometric analysis works precisely because authors write in subtly different registers \citep{grieve2023register,biber2019register}, and that formal features are theoretically meaningful carriers of narrative structure, not mere surface patterns \citep{piper2021narrative,abello2012computational}. Even under contemporary generative conditions, stylistic signals remain distinguishable. LLM-generated text retains stylometrically detectable model signatures \citep{mikros2025beyond}, though whether revision preserves or disrupts the \textit{author's} signature is a separate question. Features such as contraction, pronoun use, punctuation, and lexical variation are routinely used to characterize style, register, and authorship, making their systematic alteration by AI revision methodologically consequential.

\subsection{LLM Writing Assistance as Sociotechnical Intervention}
A parallel literature has examined detection of AI-generated text, finding that stylometric methods can distinguish human from LLM outputs \citep{przystalski2026stylometry,osullivan2025stylometric}, but that detector performance degrades with AI-revised human writing \citep{zhang2024llmcoauthor}. If LLM revision shifts the same features stylometry depends on, surface features may be altered without leaving visible traces.

The question, then, is whether these features are already being reshaped before analysis begins. At corpus scale, \citet{kobak2025delving} identify detectable lexical shifts associated with LLM-assisted writing in biomedical publications, while \citet{reinhart2025llms} show that LLM-generated text differs systematically from human writing across grammatical and rhetorical dimensions. Other work examines AI-assisted writing interactionally, showing that interface design and scaffolding shape co-writing practices and the resulting text \citep{dhillon2024shaping,lee2024design}. Most directly relevant here, \citet{Abdulhai2026distort} demonstrate that even ``minimal edits'' and ``grammar-only'' prompts can produce substantial semantic drift in AI revisions, moving toward more formal, impersonal language while sharply reducing first-person expression. 

A related strand of research has documented \textit{homogenization} effects in LLM-assisted creative work. \citet{doshi2024homogenizing} show that while LLM assistance can expand individual idea ranges, it simultaneously reduces the collective diversity of outputs across users. \citet{kremins2024homogenization} find similar patterns in creative ideation tasks, where different users produce less semantically distinct ideas when using LLM support. This phenomenon, sometimes termed \textit{register leveling}, refers to the tendency of probabilistic language models to blur distinctions between writing styles and genres, producing more uniform output regardless of input register. 

What remains less explored is whether this reshaping affects the specific features that matter for interpreting personal narratives, where irregularity, hesitation, and situated first-person expression may themselves carry evidentiary weight. Prior work on argumentative student essays and essay reformulation already suggests that LLMs alter stance, metadiscourse, and other stylistic features \citet{JIANG202517}; in personal narrative, the interpretive stakes of such reshaping may be higher.

Recent work has shown that LLM outputs can be highly sensitive to seemingly minor variations in prompt wording, format, and framing \citep{zhuo-etal-2024-prosa,ngweta-etal-2025-towards,seleznyov-etal-2025-punctuation}, with measurable variability appearing even across semantically equivalent paraphrases \citep{alhetelah-ahmad-2026-measuring,truong-etal-2025-persona}. This raises a methodological question for any study of LLM rewriting: are the observed stylistic shifts stable properties of the rewriting operation, or artifacts of a particular prompt formulation? The present study addresses this by testing three prompt conditions that vary evaluative framing while holding task constant.
\section{Methods}
\label{sec:methods}

\subsection{Dataset: EmpathicStories}

This study focuses on personal narratives rather than fiction because personal narratives claim a referential relationship to lived experience. In such texts, stylistic features like hedging, repetition, and emotional specificity are potential indexes of the narrator's relationship to the events described. The study uses the EmpathicStories corpus \citep{shen2023empathicstories}, a dataset of 1,568 first-person narratives compiled for research on narrative empathy. The corpus draws from six sources, including road-trip oral histories, Hippocorpus recollections \citep{sap2020recollection}, Reddit communities (\texttt{r/OffMyChest}, \texttt{r/CasualConversation}, and \texttt{r/TodayIAmHappy}), and Facebook MIT Confessions. This source diversity introduces natural variation in register, formality, and emotional tone.

These sources differ in several ways. Oral histories are transcribed speech, Reddit posts are informal written discourse, Hippocorpus entries are structured recollections. However, they share a common theoretical foundation as \textit{personal narrative}: each presents a first-person account of lived experience, recounting events that carry emotional, social, or moral stakes beyond mere information transfer. This shared orientation toward subjective experience positions them as a coherent category for studying how LLM rewriting affects narrative voice, even as their baseline register differences allow us to test whether normalization occurs across stylistic starting points.

Source-level variation also shapes the magnitude of normalization, a point examined in detail in Section~\ref{sec:results} and the Supplementary Materials. From the 1,112 stories meeting the minimum length criterion of 150 words, a stratified sample of $N = 300$ was drawn using a fixed random seed, with stories sampled proportionally from each data source to maintain the original source distribution. The final sample has mean length 244 words (SD = 72). Source-level checks are reported as robustness analyses below.

\subsection{Models and Prompt Conditions}

The study implements a fully crossed within-subjects design in which each narrative is rewritten under every combination of model and prompt condition. Each text therefore serves as its own control. With 300 source narratives, 3 models, and 3 prompt conditions, the design yields $300 \times 3 \times 3 = 2{,}700$ rewrites.

The study tests three frontier LLMs from major providers: OpenAI GPT-5.4 (\texttt{gpt\-5.4\-2026-02}), Anthropic Claude Sonnet 4.6 
(\texttt{claude\-sonnet\-4.6\-20260215}), and Google Gemini 3.1 Pro (\texttt{gemini\-3.1\-pro}). All API calls were made between March 1--15, 2026. These models were selected to capture variation across leading commercial systems rather than minor differences within a single provider. All models were accessed through their official APIs with temperature set to 0.7 and maximum output length set to 4,096 tokens. Temperature 0.7 balances determinism and variation; robustness checks confirm results hold across settings (see below).

To examine whether normalization depends on instruction framing, the study uses three prompt conditions:

\begin{enumerate}
    \item \textbf{Generic improvement}: \\
    \textit{``Please improve the following story while preserving its meaning.''}

    \item \textbf{Voice-preserving improvement}: \\
    \textit{``Please improve the following story while preserving the author's voice and emotional tone.''}

    \item \textbf{Rewrite-only}: \\
    \textit{``Please rewrite the following story.''}
\end{enumerate}

These conditions separate three related factors, namely broad revision framing, explicit voice-retention guidance, and the presence or absence of evaluative language. The generic condition approximates a common writing-assistant request without stylistic constraints. The voice-preserving condition tests whether a simple user-level instruction attenuates normalization. The rewrite-only condition removes the evaluative verb \textit{improve}, allowing us to assess whether normalization is tied specifically to improvement-oriented framing or emerges under rewriting more generally.

\subsection{Linguistic Markers}

The study measures normalization using 13 markers selected from computational stylistics and register studies. Selection proceeded in two stages. First, candidate features were identified from the stylometric literature \citep{stamatatos2009survey,eder2016stylometry,burrows2002delta} and register studies \citep{biber1988variation,biber2019register}. Second, features that were redundant, sparse in short personal narratives, or lacked clear interpretive payoff for questions of voice and authenticity were excluded. The resulting 13 markers span four categories.

\textbf{Function words} (2 markers) follow the Delta tradition \citep{burrows2002delta,eder2016stylometry}:

\begin{itemize}
    \item \textit{Function word ratio}: proportion of tokens that are function words.
    \item \textit{MFW coverage}: proportion of text covered by each text's 50 most frequent words.
\end{itemize}

\textbf{Vocabulary} (5 markers):

\begin{itemize}
    \item \textit{Yule's K} \citep{yule1944statistical} and \textit{Honor\'{e}'s R} \citep{honore1979some}: vocabulary concentration and hapax-based richness, both robust to text length \citep{tweedie1998variable}.
    \item \textit{MTLD} \citep{mccarthy2010mtld}: lexical diversity measure more robust to text length than traditional TTR.
    \item \textit{Mean word length} \citep{mendenhall1887characteristic}: index of vocabulary complexity.
    \item \textit{Character trigram entropy}: diversity of character-level patterns, encoding morphological habits and word boundaries while remaining topic-invariant \citep{stamatatos2009survey,stamatatos2013robustness}.
\end{itemize}

\textbf{Syntax and punctuation} (3 markers):

\begin{itemize}
    \item \textit{Mean sentence length} \citep{williams1940sentence,stamatatos2009survey}.
    \item \textit{Comma frequency} and \textit{dash frequency} \citep{stamatatos2009survey,grieve2007quantitative}: clause structure and parenthetical habits.
\end{itemize}

\textbf{Register} (3 markers) indexes formality along the written--spoken continuum \citep{biber1988variation,heylighen2002variation}:

\begin{itemize}
    \item \textit{Contraction density}: direct marker of informal register.
    \item \textit{First-person pronoun density}: narrator presence \citep{biber1988variation}.
    \item \textit{Emotion word density}: degree of explicit affective language, using the NRC Emotion Lexicon \citep{mohammad2013crowdsourcing}.
\end{itemize}

\subsection{Statistical Analysis}
For each marker, the analysis compares originals and rewrites using paired $t$-tests, with Cohen's $d$ as the standardized effect size. For stylometric features (function words, vocabulary diversity, punctuation), effect sizes were computed using pooled standard deviation to facilitate comparison with prior stylometric work; for register markers (contractions, first-person pronouns, emotion words), paired Cohen's $d$ (standard deviation of within-subject differences) was used because the within-subject design makes individual-level change the quantity of interest. Multiple comparisons across the 13 markers are controlled using Benjamini--Hochberg FDR correction at $\alpha = 0.05$. The combination of parametric tests with effect-size reporting follows standard practice in applied linguistics and computational social science \citep{dhillon2024shaping, reinhart2025llms}, where $d$ provides interpretable magnitude estimates and facilitates cross-study comparison.

The analysis reports both marker-level and dimension-level results and compares effect sizes across prompt conditions to assess whether voice-preservation instructions mitigate normalization. To assess robustness, four validation analyses were conducted on subsamples:

\paragraph{Self-consistency.} Three independent rewriting passes on 50 stories yielded median ICC = 0.83 across markers. Most markers exceeded 0.90; function word ratio (0.87) and MTLD (0.82) showed moderate reliability, while connective density was lower (0.66). The largest effects reported below occur on high-reliability markers.

\paragraph{Protocol sensitivity.} To test whether findings depend on message placement, prompts presented in system versus user messages were compared for 100 stories. For each marker, Cohen's $d$ was computed under both protocols and these effect-size vectors were correlated across markers. The median correlation was $r = 0.82$, indicating that the pattern of which markers change and by how much is largely stable across prompt-delivery formats.

\paragraph{Temperature stability.} Rewriting was repeated at temperatures 0.0, 0.7, and 1.0 on 50 stories. For each temperature setting, Cohen's $d$ was computed for each marker and these effect-size vectors were correlated across settings. Pairwise correlations exceeded $r = 0.96$ for all models (GPT-5.4: $r = 0.99$; Claude Sonnet: $r = 0.99$; Gemini: $r = 0.97$), indicating that the main transformation patterns are stable across stochasticity settings. Higher temperature increases output variability but does not alter the directional tendencies of normalization.

\paragraph{Source stratification.} Supplementary analyses test whether the main directional effects hold across source categories, given that baseline register varies across the corpus. Full source-level breakdowns are provided in the Supplementary Materials (Section 4).

\paragraph{Semantic preservation.} To confirm that rewrites alter form rather than content, semantic similarity was computed using sentence embeddings (all-MiniLM-L6-v2). Mean cosine similarity between originals and rewrites was 0.90 for GPT-5.4, 0.85 for Claude, and 0.84 for Gemini---substantially higher than random text pairs (0.22) or same-source pairs (0.28). A weak negative correlation between stylistic change magnitude and semantic similarity ($r = -0.26$) indicates that more aggressive rewrites slightly reduce meaning preservation, but even high-change texts maintain $>$83\% similarity. This confirms that the transformations documented below operate primarily on stylistic form rather than semantic content.
\section{Results}
\label{sec:results}

The analysis covers 2,700 rewrites derived from 300 source narratives rewritten by 3 models under 3 prompt conditions. Change is measured using 13 markers grounded in computational stylistics and register studies (described in Methods).

\subsection{Directional Convergence Across Models}

All three models systematically alter the linguistic properties of personal narratives when asked to revise them. Under the baseline \textit{generic improvement} prompt, all 39 marker--model combinations (13 markers $\times$ 3 models) are statistically significant after FDR correction ($\alpha = 0.05$), with mean absolute effect sizes in the large range (Table~\ref{tab:overview}). Because the marker set was selected based on prior stylistic and psycholinguistic research rather than exhaustively derived, this convergence reflects shared behavior within that operationalization rather than an open-ended discovery.

\begin{table}[!htbp]
\centering
\begin{tabular}{lcc}
\toprule
\textbf{Model} & \textbf{Sig. Markers} & \textbf{Mean $|d|$} \\
\midrule
GPT-5.4 & 13/13 (100\%) & 0.60 \\
Claude Sonnet 4.6 & 13/13 (100\%) & 1.29 \\
Gemini 3.1 Pro & 13/13 (100\%) & 1.45 \\
\bottomrule
\end{tabular}
\caption{Marker-level change under the generic improvement prompt. Mean $|d|$ denotes the unweighted average of absolute Cohen's $d$ values across the 13 markers. Each marker's $d$ was computed from paired comparisons of 300 original--rewrite text pairs. For reference, Cohen's conventional benchmarks are small ($d = 0.2$), medium ($d = 0.5$), and large ($d = 0.8$).}
\label{tab:overview}
\end{table}

The magnitude of these effects is substantial: 59\% of marker--model combinations exceed the conventional threshold for ``large'' effects ($|d| > 0.8$). Crucially, all 13 markers show consistent directional effects across models, with no marker moving in opposite directions for different models. Despite different architectures, training data, and providers, all three models push personal narratives in the same direction: toward texts that are more lexically varied, less contraction-heavy, less first-person-centered, and more punctuationally elaborated.

\FloatBarrier
\subsection{Normalization Patterns}

Table~\ref{tab:markers} presents effect sizes for all 13 markers under the generic improvement prompt, organized by the analytical tradition from which each derives.

\begin{table*}[!htbp]
\centering
\begin{tabular}{llccc}
\toprule
\textbf{Category} & \textbf{Marker} & \textbf{$d$ (M)} & \textbf{Direction} & \textbf{\% Change} \\
\midrule
Function Words & MFW coverage & $-1.76$ & $\downarrow$ & $-11.5\%$ \\
& Function word ratio & $-1.13$ & $\downarrow$ & $-8.4\%$ \\
\midrule
Vocabulary & Honor\'{e}'s R & $+1.74$ & $\uparrow$ & $+32.7\%$ \\
& MTLD & $+1.59$ & $\uparrow$ & $+52.7\%$ \\
& Word length & $+1.50$ & $\uparrow$ & $+8.5\%$ \\
& Character trigrams & $+1.37$ & $\uparrow$ & $+3.2\%$ \\
& Yule's K & $-0.91$ & $\downarrow$ & $-19.9\%$ \\
\midrule
Syntax \& Punct. & Comma frequency & $+1.40$ & $\uparrow$ & $+66.6\%$ \\
& Dash frequency & $+1.12$ & $\uparrow$ & $+325.7\%$ \\
& Sentence length & $+0.33$ & $\uparrow$ & $+7.1\%$ \\
\midrule
Register & Emotion words & $+0.58$ & $\uparrow$ & $+14.4\%$ \\
& First-person pronouns & $-0.57$ & $\downarrow$ & $-8.6\%$ \\
& Contractions & $-0.48$ & $\downarrow$ & $-31.4\%$ \\
\bottomrule
\end{tabular}
\caption{Effect sizes for 13 markers under the generic improvement prompt, organized by analytical tradition. $d$ (M) = mean Cohen's $d$ across three models. Direction arrows indicate whether rewrites increase ($\uparrow$) or decrease ($\downarrow$) the marker value. All markers show consistent direction across models.}
\label{tab:markers}
\end{table*}

\begin{figure*}[!htbp]
\centering
\includegraphics[width=\textwidth]{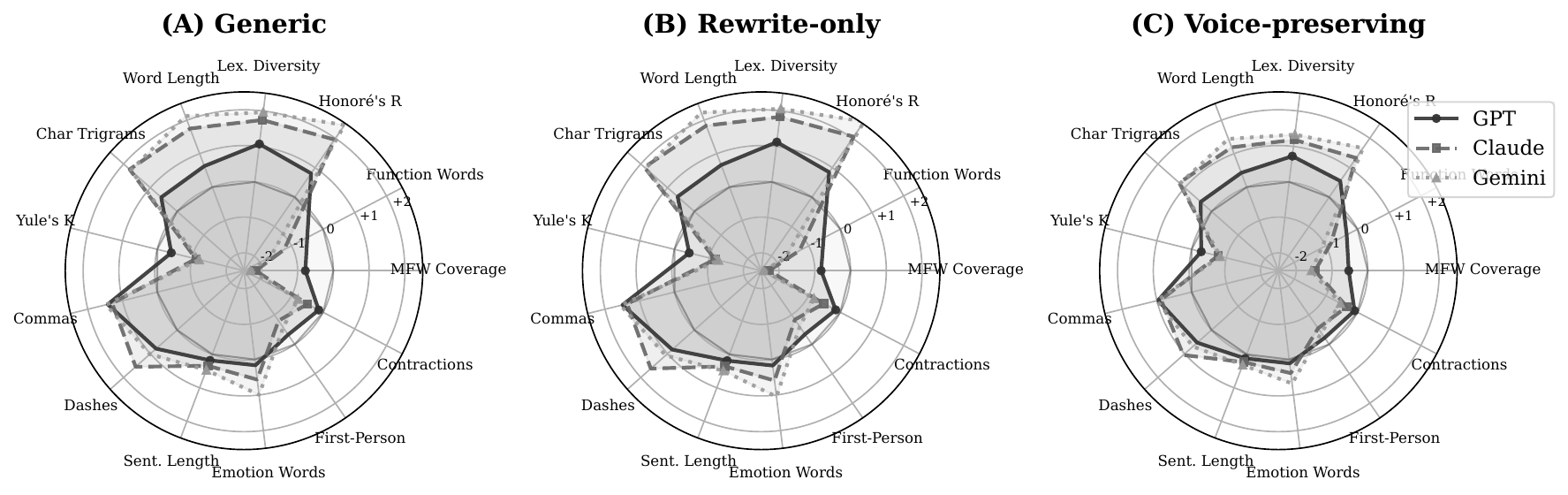}
\caption{Radar plots showing normalization pattern across 13 markers for each prompt condition. (A) Generic improvement, (B) Rewrite-only, (C) Voice-preserving. Radial distance from center indicates signed Cohen's $d$ (positive = increase, negative = decrease). All three models show the same directional pattern across all conditions: the ``fingerprint'' of normalization is remarkably consistent regardless of prompt framing. Voice-preserving instructions (C) reduce effect magnitude but preserve the overall shape.}
\label{fig:radar}
\textit{Alt text:} Three radar charts showing identical spiky polygon shapes across panels, with vertices extending outward on vocabulary measures and inward on function word measures.
\end{figure*}

Figure~\ref{fig:radar} visualizes the normalization pattern as a three-panel radar plot, one for each prompt condition. Panel (A) shows the generic improvement condition. The shape of each model's profile is similar: function words and first-person pronouns decrease; vocabulary diversity, word length, and punctuation elaboration increase. Panels (B) and (C) are discussed in Section~\ref{sec:inherent}.

LLM rewriting draws personal narratives toward a common register of polished, mediated prose. The pattern is best understood through two primary dimensions of change:

\begin{itemize}
    \item \textit{Register formalization}: Contraction density decreases by 31\% ($d = -0.48$), and first-person pronoun density decreases by 9\% ($d = -0.57$). Mean word length increases by 8.5\% ($d = +1.50$), while vocabulary diversity measures rise substantially (MTLD +53\%, Honor\'{e}'s R +33\%). The vocabulary changes are consistent with this interpretation: formal register favors richer, less repetitive lexical choices and proportionally more content words relative to function words (MFW coverage $-11.5\%$, function word ratio $-8.4\%$).

    \item \textit{Punctuation-mediated editorial smoothing}: Comma frequency increases by 67\% ($d = +1.40$), from approximately 8 to 13 commas per text. Dash frequency increases by 326\% ($d = +1.12$). While absolute counts remain low, the proportion of texts containing at least one dash rises from 26\% to 72\%. These changes indicate systematic introduction of parenthetical structures and syntactic segmentation.
\end{itemize}

The two dimensions work together: register formalization affects lexical choice and firts-person pronoun density, while punctuation elaboration reshapes the rhythmic and syntactic texture. Together, these shifts indicate movement toward more formal, elaborated expression, a register associated with edited prose rather than spontaneous personal narration. Beyond lexical and punctuation changes, rewriting also reorganizes sentence structure: GPT-5.4 and Gemini reduce mean sentence counts by 4--5, merging clauses into longer units, while Claude increases them by 9, splitting constructions into discrete statements.

The strongest effects occur on markers of discourse presentation: contractions, punctuation patterns, clause structure, and vocabulary distribution. These are the features that give a narrative its manner of telling, whether spontaneous or composed, speech-like or writerly, immediate or retrospective. The largest effect sizes occur on function words and vocabulary measures (features central to stylometric methods; \citealt{stamatatos2009survey,burrows2002delta}), suggesting that normalization reshapes the low-level stylistic texture that distinguishes individual texts.

Figure~\ref{fig:normalization}, Panel (A) provides another view of this pattern, showing the direction of change for each marker sorted by effect size. The largest effects cluster around function words and vocabulary diversity measures, which bear directly on register formalization; punctuation markers (comma and dash frequency) show large effects in the elaborative direction.

To ensure these shifts represent generalized normalization rather than simple genre translation, results were stratified by data source. Twelve of thirteen markers showed consistent directional shifts across Hippocorpus recollections, oral-history transcripts, and Reddit posts (Supplementary Table~S2). Effect magnitudes vary, with oral-history transcripts showing the largest effects on contraction density ($d = -1.15$), Reddit narratives showing intermediate effects ($d = -0.53$), and Hippocorpus recollections showing the smallest ($d = -0.11$). The one exception to directional consistency is contraction density itself, which decreases in oral transcripts but slightly increases in already-formal recollections, suggesting normalization toward a common register from both directions.

\FloatBarrier
\subsection{Normalization Persists Across Prompt Conditions}
\label{sec:inherent}

The next question is whether normalization depends specifically on evaluative framing, that is, whether saying ``improve'' triggers the observed transformations. To address this, the analysis compares the \textit{generic improvement} condition against the \textit{rewrite-only} condition, which removes the evaluative verb entirely, and the \textit{voice-preserving} condition, which explicitly instructs the model to retain authorial voice.

\begin{table}[!htbp]
\centering
\small
\begin{tabular}{lccc}
\toprule
\textbf{Condition} & \textbf{$|d|$} & \textbf{vs.\ Gen.} & \textbf{Dir.\ Agr.} \\
\midrule
Generic & 1.11 & --- & --- \\
Rewrite-only & 1.16 & $+5\%$ & 100\% \\
Voice-preserving & 0.76 & $-32\%$ & 85\% \\
\bottomrule
\end{tabular}
\caption{Comparison across prompt conditions. $|d|$ = mean absolute Cohen's $d$ across 13 markers and 3 models; each marker's $d$ is computed from 300 paired original--rewrite comparisons, then averaged. Comparisons of these aggregated values are descriptive summaries of effect-size magnitude rather than formal inferential tests. Dir.\ Agr.\ = percentage of markers moving in the same direction as the generic condition.}
\label{tab:prompt-comparison}
\end{table}

The \textit{rewrite-only} condition produces nearly the same overall pattern as \textit{generic improvement}: mean absolute effect sizes are comparable ($|d| = 1.16$ vs.\ $1.11$), and all 13 markers move in the same direction (Table~\ref{tab:prompt-comparison}). Removing overtly evaluative language does not reduce the tendency of models to regularize the texts they rewrite. While this comparison does not identify the source of that tendency, it does show that the observed transformations are not reducible to the wording of a single improvement-oriented prompt. 

Voice-preserving instructions reduce normalization, though only partially. Across models, mean absolute effect size falls from $|d| = 1.11$ to $|d| = 0.76$, a reduction of 32\% (Table~\ref{tab:voice-effect}). Direction agreement drops to 85\%, with two markers flipping for GPT and Claude: contraction density reverses from decreasing to slightly increasing, and mean sentence length reverses from increasing to decreasing---suggesting partial overcorrection on these features. At the same time, the residual mean absolute effect ($|d| = 0.76$) still approaches the conventional threshold for ``large'' effects. Vocabulary inflation continues (MTLD still increases by 35\% under voice-preserving prompts), and function word reduction persists (MFW coverage still decreases by 8\%).

\begin{table}[!htbp]
\centering
\small
\begin{tabular}{lcccc}
\toprule
\textbf{Model} & \textbf{Gen.} & \textbf{Rew.} & \textbf{Voice} & \textbf{Red.} \\
\midrule
GPT-5.4 & 0.60 & 0.78 & 0.37 & 38\% \\
Claude Sonnet & 1.29 & 1.15 & 0.93 & 28\% \\
Gemini 3.1 Pro & 1.45 & 1.57 & 0.97 & 33\% \\
\midrule
\textbf{Mean} & 1.11 & 1.16 & 0.76 & \textbf{32\%} \\
\bottomrule
\end{tabular}
\caption{Mean $|d|$ by model and prompt condition. Gen.\ = generic improvement; Rew.\ = rewrite-only; Voice = voice-preserving; Red.\ = reduction from generic to voice-preserving.}
\label{tab:voice-effect}
\end{table}

While the broad normalization pattern is shared, the three models differ substantially in degree. GPT-5.4 shows the smallest overall effect sizes (mean $|d| = 0.60$), with notably smaller effects on vocabulary inflation and function word reduction. Claude Sonnet 4.6 produces intermediate effects (mean $|d| = 1.29$), while Gemini 3.1 Pro produces the largest overall transformations (mean $|d| = 1.45$), including the strongest increases in lexical diversity ($d = 1.96$) and vocabulary richness ($d = 2.45$). The same editing instruction produces different magnitudes of transformation depending on the model used.

\begin{figure*}[!htbp]
\centering
\includegraphics[width=\textwidth]{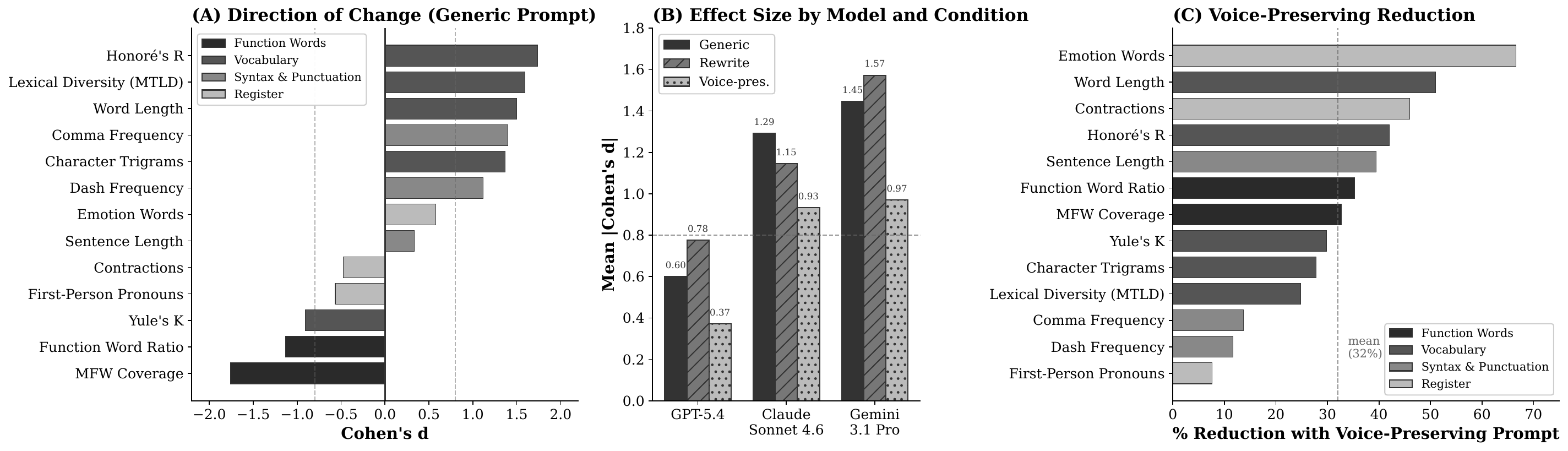}
\caption{Normalization effects across 13 linguistic markers. (A) Signed Cohen's $d$ under generic improvement prompt; positive = increase, negative = decrease. Grayscale indicates marker category. Dashed lines mark $|d| = 0.8$. (B) Mean $|d|$ by model and prompt condition. (C) Percentage reduction in $|d|$ under voice-preserving versus generic prompts.}
\label{fig:normalization}
\textit{Alt text:} Panel A shows bars diverging left and right from zero. Panel B shows grouped bars decreasing across conditions. Panel C shows horizontal bars of varying lengths.
\end{figure*}

Figure~\ref{fig:normalization}, Panels (B) and (C) visualize these findings. Panel (B) shows that generic and rewrite-only produce nearly identical effects, while voice-preserving shows smaller effects across all models. Panel (C) shows the percentage reduction achieved by voice-preserving versus generic instructions for each marker, revealing that some features respond well to explicit preservation prompts (emotion words, word length) while others remain resistant (first-person pronouns, function words).

\subsubsection{Differential Response to Voice-Preserving Prompts}
Panel~C shows substantial variation in how individual markers respond to voice-preserving instructions. Emotion words and word length show the largest reductions (approximately 55\%), while first-person pronouns and function word ratio show the smallest (approximately 10--15\%). Between these extremes, contraction density shows an intermediate pattern, shifting from $d = -0.50$ under generic prompts toward $d \approx 0$ under voice-preserving prompts.

One possible interpretation is that voice-preserving prompts more effectively suppress elaborative tendencies (i.e., changes that add or expand material) than they preserve features the model tends to remove or compress. Under this reading, voice-preserving outputs would be less embellished than generic rewrites, but still transformed from the original in ways that resist explicit instruction. Model-specific anomalies, including a case where voice-preserving instructions appear to intensify rather than attenuate homogenization, are examined in Supplementary Materials.

\FloatBarrier

\FloatBarrier
\subsection{Stylometric Convergence and Source-Text Recoverability}

The preceding sections established that LLM rewriting produces large, directionally consistent shifts across 13 markers. Because the largest effects occur on function-word and vocabulary measures---features widely used in stylometric methods to distinguish texts---the next question is whether these same changes reorganize stylometric space and affect stylometric methods in practice.

To assess the practical consequences of normalization for computational stylometry, pairwise distances were computed and texts projected into a two-dimensional stylometric space using PCA on a feature set combining character n-gram measures with word length. Given the short average length of the samples in this corpus (244 words), traditional word-frequency Delta was not used, as it is generally less well suited to short texts \citep{eder2015does}. Instead, the analysis used character n-gram–based measures—specifically bigram, trigram, and quadrigram entropy and hapax ratios—which build on approaches shown to be robust for short and heterogeneous texts in authorship analysis \citep{stamatatos2013robustness}, supplemented by mean word length.

\begin{figure*}[!htbp]
\centering
\includegraphics[width=\textwidth]{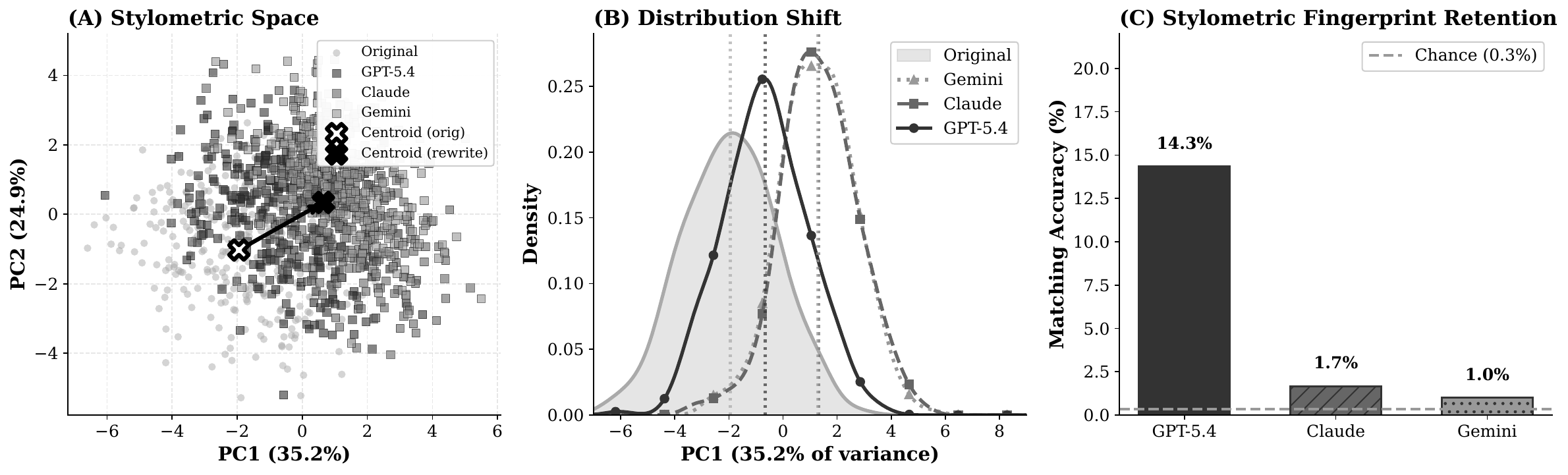}
\caption{Stylometric convergence under LLM rewriting. (A) PCA projection based on character n-gram measures and word length, showing originals (gray circles) scattered widely versus rewrites (squares) clustering tightly. Arrow indicates centroid shift from original (white X) to rewrite (black X). (B) Distribution shift along PC1: originals (dashed gray) show broad variation. (C) Source recoverability: using nearest-neighbor matching to test whether rewrites can be matched back to their source texts under this feature representation, GPT-5.4 rewrites achieve 14.3\% matching accuracy (vs.\ 0.3\% chance), while Claude (1.7\%) and Gemini (1.0\%) rewrites fall to near-chance levels.}
\label{fig:convergence}
\textit{Alt text:} Panel A shows scattered gray circles with tightly clustered colored squares near center. Panel B shows one broad curve versus three narrow peaked curves. Panel C shows three bars of decreasing height.
\end{figure*}

Figure~\ref{fig:convergence} visualizes the results. Panel (A) shows a PCA projection of the stylometric feature space. Original texts (gray circles) are more broadly dispersed, whereas rewrites (colored squares) cluster more tightly. The first two principal components explain 35.2\% and 24.9\% of the variance, respectively; PC1 loads primarily on character n-gram entropy and word length. Despite differences in architecture and training, all three models move rewrites toward a similar region of the space. Panel (B) shows the corresponding shift along PC1: the original texts (dashed line, gray fill) exhibit a broader distribution centered near zero, while all three models produce distributions that are both shifted and narrower, with Gemini showing the largest displacement and GPT the smallest.

A further test examined whether stylometric features could be used to match rewritten texts back to their source texts. For each rewrite, the distance to all 300 source texts was computed using the same feature set, and checked whether its nearest neighbor was in fact the original. While this is not traditional multi-author attribution, which would require multiple texts per author, it does test whether rewriting preserves enough stylometric signal for a text to remain identifiable. Under the generic prompt, GPT-5.4 rewrites were matched to their source texts 14.3\% of the time, compared with a 0.3\% chance baseline, whereas Claude Sonnet and Gemini 3.1 Pro dropped to 1.7\% and 1.0\%, respectively. Even the best-performing model leaves over 85\% of texts stylometrically unrecoverable

To assess whether this pattern was specific to character-level analysis, the procedure was repeated using traditional word-based Burrows' Delta over Most Frequent Words. Although Delta is more commonly used on longer texts, recent work has applied it to AI–human comparisons on texts of similar length to capture aggregate stylistic differences \citep{osullivan2025stylometric}. The pattern was similar: GPT-5.4 reached 15.7\% matching accuracy, compared with 2.3\% for Claude and 0.3\% for Gemini (Supplementary Table~S3). The convergence of results across character-level and word-level feature spaces suggests that the difference lies less in the type of representation than in the magnitude of stylometric compression: GPT-5.4 preserves more recoverable source signal, whereas Claude and especially Gemini compress texts more strongly toward a common profile.

\subsubsection{Formal Variance Compression}

The visual convergence documented above can be quantified through formal variance analysis. Using Levene's test for equality of variances across 58 stylometric features (combining the 13 core markers with additional character n-gram and lexical measures), Claude Sonnet reduced variance in 78\% of features (45/58) with a median reduction of 26\% ($p < 0.05$ for 72\% of features). Multivariate dispersion, measured as mean Euclidean distance from the group centroid in standardized feature space, decreased by 9\% (Cohen's $d = 0.36$, Mann-Whitney $U$, $p < 0.0001$). GPT-5.4 showed smaller compression (60\% of features), while Gemini 3.1 Pro approached Claude's level (74\%).

This finding sharpens the methodological point. Rewriting shifts features in a common direction across models and reduces dispersion between texts. Even under prompts that ostensibly preserve voice, stylometric variance decreases. A perplexity check using GPT-2 as reference model corroborates this convergence: all three systems produce rewrites with lower perplexity than originals (GPT-5.4 $d = 1.18$; Claude $d = 0.44$; Gemini $d = 0.19$), indicating that vocabulary diversifies while its distribution becomes more predictable. For corpus-based research that depends on textual heterogeneity, this compression complicates the interpretive foundations of computational stylistics.

\subsection{Narrative Stance Shift}
\label{sec:stance}

The stylometric convergence documented above raises a further question: if rewrites cluster together in character n-gram space, do they also converge in how narrators position themselves toward their own experience? The register shifts already documented, such as contractions expanding and first-person density falling, suggest changes not only to surface form but also to narrative stance. The difference between ``I can't stop crying'' and ``I cannot stop crying'' involves a shift in immediacy and self-positioning. To examine this dimension more systematically, five markers were computed drawing on concepts from narrative theory \citep{labov1967narrative,bruner1991narrative} and discourse analysis.

\begin{figure}[!htbp]
\centering
\includegraphics[width=0.75\columnwidth]{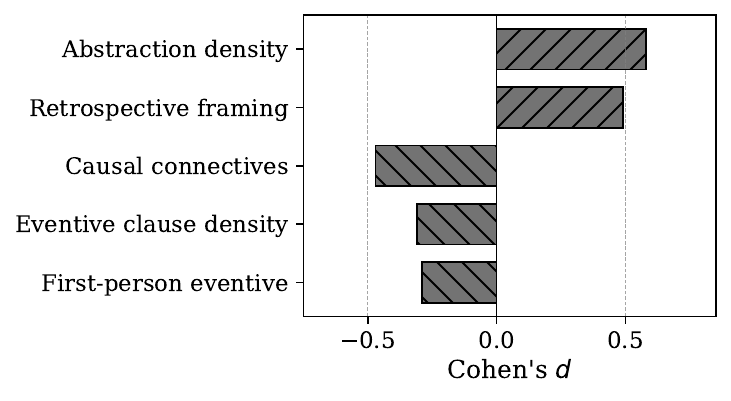}
\caption{Narrative stance shift under LLM rewriting. Bars show signed Cohen's $d$ for five stance markers; positive values indicate features that increase in rewrites, negative values indicate decreases.}
\label{fig:stance}
\textit{Alt text:} Five horizontal bars diverging from a central zero line. Two bars extend right, three extend left, with uniform gray shading throughout.
\end{figure}

Figure~\ref{fig:stance} summarizes effect sizes for these five markers under the generic improvement prompt (marker definitions appear in Supplementary Materials). The results reveal two linked shifts in how narrators position themselves relative to their experience.

Eventive clause density (proportion of clauses containing action or perception verbs; $d = -0.31$) and first-person eventive constructions (``I walked,'' ``I saw''; $d = -0.29$) both decrease, while retrospective framing (``looking back,'' ``I now realize''; $d = +0.49$) increases. This is a move from embedded narration to distanced retrospection. The narrator appears less inside unfolding events---describing what happened as it happened---and more outside of them, looking back with retrospective understanding.

Causal connectives (tokens like ``because'' and ``therefore,'' and phrases like ``as a result''; $d = -0.47$) decrease, while abstraction density (abstract nouns like ``situation'', ``realization''; $d = +0.58$) increases. This is a move from explicit reasoning to compressed interpretive abstraction, suggesting that LLM rewriting replaces one mode of explanation with another. The narrator who emerges from AI revision is one who has already understood, whose backward-looking wisdom is presumed rather than argued through causal reasoning. These two shifts are linked but distinct. The first concerns where the narrator stands relative to events; the second concerns how understanding is conveyed. Together, these shifts indicate a movement across the corpus from situated, eventive narration toward more retrospective and abstract narration.

Voice-preserving prompts reduce some of these effects more than others. Eventive clause density, which drops under generic prompts ($d = -0.31$), shows almost no change under voice-preserving prompts ($d = -0.01$), a 97\% reduction in effect size. Abstraction density effects are reduced by 76\%. However, retrospective framing continues to increase and causal connectives continue to decrease even under voice-preserving conditions, suggesting that the shift in explanatory style is harder to suppress through prompting than the shift in narrated positioning.

The stance-shift findings complement the stylometric results reported above. Where the stylometric analysis shows that rewrites converge in character n-gram space, the stance analysis shows a parallel convergence in narrative mode. Texts move from enacted, causally-argued narration toward distanced, abstractly-framed interpretation. Both analyses point to an underlying phenomenon of selective normalization, but capture different dimensions of its operation.

\subsection{Qualitative Examples}

The quantitative results show that LLM rewrites consistently reduce function-word density, contraction use, and first-person presence while increasing lexical diversity, word length, and punctuation elaboration. The qualitative examples below trace how that change becomes legible at the level of individual narratives Examples were selected for high effect magnitudes across the three focal mechanisms.\footnote{Italics in all quoted passages are added to highlight features under discussion.}

\subsubsection{Example 1: Register Formalization}

Across the corpus, contraction density and function-word ratio both decline under LLM revision. This case shows how those aggregate shifts manifest as register elevation in practice.

\begin{quote}
\textbf{Original:} ``\textit{We'd} been friends since year 9 and stayed friends up until last year when we decided to date, the year we finished school. We ended on what I think is good terms as we broke up mutually over the phone agreeing that we have a \textit{reoccurring} [sic] issue we argue over that we \textit{don't} think we have the maturity to solve but \textit{I'm} absolutely broken. As \textit{I've} left school, I \textit{don't} have many friends and my boyfriend was the main person I talk to everyday. \textit{We'd} only been dating for nearly a year and a half but it still hurts so much. [...] \textit{I'm} so lost and hurt and \textit{I can't} stop crying.''
\end{quote}

\begin{quote}
\textbf{GPT-5.4 (generic improvement):} ``\textit{We had} been friends since Year 9, and we stayed close until last year, the year we finished school, when we decided to start dating. We broke up on what I think were good terms. It was a mutual decision, made over the phone, because we both knew there was a \textit{recurring} issue we kept arguing about and neither of us felt mature enough to fix it. Even so, \textit{I am} absolutely broken. Since leaving school, \textit{I have not} had many friends, and my boyfriend was the main person I talked to every day. \textit{We were} only together for nearly a year and a half, but it still hurts so much. [...] \textit{I feel} so lost and hurt, and \textit{I cannot} stop crying.''
\end{quote}

The rewrite expands contractions, corrects spelling, and reorganizes a run-on sequence into discrete clauses with explicit connectives. In doing so, it alters the affective texture of the passage. The original's breathless accumulation conveys overwhelm through syntax; the revision recasts the same urgency as composed, sequential reflection. ``I can't stop crying'' becomes ``I cannot stop crying''---while semantically equivalent, this is registrally more formal and less immediate. The result is a shift from enacted to summarized experience.

\subsubsection{Example 2: Vocabulary Inflation}

MTLD and Honor\'{e}'s $R$ increase substantially across the corpus. This case shows that vocabulary inflation occurs even under the rewrite-only prompt.

\begin{quote}
\textbf{Original:} ``I'm the chief financial officer as well as the chief actuary here at the Geisman Health Plan. So what we primarily do here is we sell health insurance products. One of my primary activities as the chief actuary is to price those insurance products. We have to make projections on what healthcare costs are \textit{gonna} be for the upcoming year. It's one of those professions, because of the challenge of getting through the process, you really develop some unique skills. But in any field you go into, try to go toward an area where you can have a broad perspective. [...] I \textit{wouldn't just shut yourself off} to those opportunities. Those growth opportunities are so important in life.''
\end{quote}

\begin{quote}
\textbf{Gemini (rewrite-only):} ``As the Chief Financial Officer and Chief Actuary at Geisman Health Plan, my primary focus is our core business: providing health insurance. A major part of my actuarial role involves pricing these products, which requires \textit{accurately forecasting} healthcare costs for the upcoming year. \textit{Navigating the rigorous demands} of this profession naturally equips you with a \textit{highly specialized and unique} skill set. Regardless of the field you choose, I always advise seeking out roles that offer a broad perspective. [...] I \textit{urge you never to close yourself off to the unknown}. Leaning into those unexpected opportunities is ultimately what drives the most important growth in our lives.''
\end{quote}

In this rewrite, the colloquial ``gonna'' disappears, simple phrasing expands into elaborated nominal constructions, and the oral texture of the original gives way to polished advisory prose. Stylometrically, repeated lexical habits are redistributed into a more varied and less speaker-specific vocabulary profile.

\subsubsection{Example 3: Stylization}

This example illustrates the counterintuitive finding that voice-preserving instructions can intensify stylization instead of preventing it.

\begin{quote}
\textbf{Original:} ``While I was scrolling through the server, I noticed the one person I didn't know had changed their name to `disgusting trash' and I became worried, not because I knew them, but because I genuinely love to keep everyone around me happy, including strangers. Not knowing this person and never having spoken to them, I message them and ask them if they're okay and talk for a little bit. [...] For the next few minutes I felt nothing but the purest form of happiness as I realised that I just made someone's day better!''
\end{quote}

\begin{quote}
\textbf{Claude (voice-preserving):} ``While scrolling through the server one evening, I noticed that this stranger had changed their display name to `disgusting trash.' \textit{Something inside me shifted.} It wasn't concern born from friendship or obligation---I didn't know this person at all---but I've always carried this \textit{deep, almost instinctive need} to see the people around me doing okay. Even strangers. Maybe especially strangers. So I reached out. [...] For the next few minutes, I felt something I can only describe as \textit{pure}. Not proud, not accomplished---just genuinely, quietly happy. \textit{The kind of happiness that settles over you softly} when you realize that a small, unplanned moment of kindness was somehow exactly what another person needed to hear.''
\end{quote}

Despite explicit instructions to preserve voice, Claude reimagines the passage as introspective literary prose, introducing interior metaphor, sensory imagery, and elaborated affect. The model supplies a more conventionally expressive persona than the narrator originally performed. The instruction to preserve voice appears to function less as a constraint than as a license to render voice more emotionally legible, illustrating the quantitative finding that stylometric compression can increase under voice-preserving prompts.

\section{Discussion}
\label{sec:discussion}

Across three providers, three prompt conditions, and 13 markers, LLM rewriting decreases function words and first-person pronouns, while increasing vocabulary diversity, punctuation, and word length. Voice-preserving instructions reduce effect magnitude by about one-third but preserve direction. The largest effects occur on function words and vocabulary measures, features central to computational stylometry \citep{stamatatos2009survey,burrows2002delta}. 

Source-stratified analyses show that normalization is sensitive to the starting register of the source text. Informal Reddit narratives shift more strongly than Hippocorpus recollections, which already lie closer to the models’ preferred register, while oral-history transcripts show the strongest effects of all. Contraction density provides a useful illustration: it decreases in oral transcripts but increases slightly in already-formal recollections, suggesting calibration toward a common register rather than simple suppression of informality.

At a broader level, these shifts move personal narratives away from interpersonal narration and toward a more document-like written register. The markers most affected by LLM rewriting—contraction, discourse marking, exclamation, paratactic sequencing, and first-person immediacy—help sustain a speech-like, situated mode of telling. Qualitatively, hesitation gives way to coherence, repetition to variation, and colloquial phrasing to more standardized expression. Sentence structure shifts as well, with models moving texts away from paratactic accumulation and toward more composed prose.

The prompt-comparison results further suggest that user control is asymmetric. Voice-preserving prompts curb some additive tendencies more effectively than they preserve the original distribution of stylistic features. Users appear to have more leverage over what the model adds than over what it removes. Embellishing tendencies such as sentence lengthening, sentiment amplification, or lexical inflation respond more readily to explicit instruction than subtractive tendencies such as the erosion of contraction patterns, first-person presence, or function-word distributions. Careful prompting may therefore curb textual flourish without reliably preserving the stylistic residue of lived narration.

\subsection{LLM Rewriting as Textual Mediation}

The present findings add empirical support to emerging work on LLM-based writing assistance as a form of textual mediation. Unlike transcription, OCR, or other conversion-based processes, LLM rewriting operates through fluent rephrasing at the level of wording, register, and narrative texture, often without leaving clear procedural traces. This is especially consequential for personal narrative and other speech-derived materials, where textual features are more readily read as traces of a situated speaker or lived perspective. It therefore makes textual cues harder to interpret, because features that still appear to register voice or lived perspective may already have been reshaped by model-mediated revision.

These implications extend to the social conditions of textual production and analysis. Once AI-assisted editing becomes expected, writers may begin composing in anticipation of it. Markers of immediacy or sincerity may become increasingly strategic, while models in turn learn to preserve, suppress, or simulate those same effects. This transforms textual analysis as well, since close reading often depends on treating such features as evidence of stance, immediacy, difficulty, or emotional weight. Such inferences depend on the assumption that textual form bears a sufficiently stable relation to conditions of production. When those features may have been preserved, suppressed, or strategically reintroduced in anticipation of, or response to, model revision, that inferential chain becomes less secure.

The stylometric analyses extend this concern from the level of individual texts to that of the corpus. In the PCA projection, originals occupy a broad cloud, while rewrites cluster much more tightly. Levene’s test across 58 stylometric features corroborates this compression: Claude Sonnet reduced variance in 78\% of features, with a median reduction of 26\%. Perplexity provides independent support as well, with all three models producing more predictable prose than the originals. Vocabulary diversity increases, but the distribution of that vocabulary becomes less text-specific. If AI-assisted revision becomes common in contemporary archives, the result may be reduced between-text variance overall. That matters for DH practices that depend on stylistic spread rather than just central tendency, including clustering, typology, anomaly detection, and comparative corpus analysis.

\subsection{Implications for Digital Humanities}

A practical question for researchers and archivists is whether these effects can be mitigated through prompting. The results suggest that prompting helps, though only partially. Removing the word ``improve'' changes little about the broad direction or magnitude of transformation, while voice-preserving instructions reduce effect sizes by about one-third but leave substantial change in place. Model differences reinforce this point. GPT-5.4 shows the smallest average effect sizes and produces rewrites that remain more recoverable to their source texts under the matching procedure, while Claude Sonnet 4.6 and Gemini 3.1 Pro produce much stronger compression and near-chance matching accuracy under generic prompting. The variation is therefore one of magnitude as well as direction, rendering model choice a methodological variable in its own right.

In terms of DH project design, the most immediate recommendation is provenance metadata: recording whether texts have been revised, cleaned, or transcribed with AI assistance, and where possible specifying which tools were used. For projects using TEI or other structured metadata, explicit AI-assistance flags would allow researchers to filter or stratify by revision status. This documentation imperative is especially urgent for project types in which textual form is itself part of the evidence. Oral-history digitization initiatives that rely on AI-assisted transcription risk introducing normalization at the very point where speech features are being converted to text. Social-media archives may also capture LLM-revised posts without distinguishing them from unrevised content. In such cases, downstream analyses may be compromised if AI mediation remains undocumented.

\subsection{Limitations and Future Directions}

The scope of the claims advanced here is bounded by several limitations. First, the present design does not isolate the source of the observed effects. It cannot distinguish among possible contributions from pretraining data, instruction tuning, alignment procedures, system-level defaults, or interactions among these factors. Second, while the patterns reported are robust across models and prompt conditions, the study is confined to a single English-language corpus and does not establish cross-genre, cross-linguistic, or cross-cultural generalizability. Third, the texts in the corpus average 244 words, shorter than the 2,500--5,000 words typically recommended for word-based authorship attribution using Burrows' Delta \citep{eder2015does}. To address this, the stylometric convergence analysis uses character n-gram features, which are methodologically appropriate for short texts \citep{stamatatos2013robustness}. The source-rewrite matching analysis is not traditional multi-author attribution but a test of whether rewrites can be matched back to their unique source texts under a particular feature representation; the convergence findings should therefore be understood as aggregate patterns across 300 texts rather than reliable single-text diagnostics. More fundamentally, it remains unclear which of the shifts detected by the markers used here are perceptible to human readers, or which matter most for interpretation. The measures employed are grounded in prior stylistic and psycholinguistic research, but they remain proxy features associated with voice rather than direct observations of voice as a lived or readerly phenomenon.

These limitations point to several directions for future work. One is triangulation with reader-based studies to examine whether the changes identified here are perceptible and interpretively salient to human readers. A second is expansion across genres and languages, especially in corpora where orality, dialect, or non-standard register are central. A third is methodological: the field would benefit from tools for detecting AI-mediated text, or at least for enriching metadata around AI-assisted revision, so that researchers can better assess the interpretive status of contemporary corpora. Finally, future work should investigate more directly which components of current LLM pipelines contribute to normalization, since the present study identifies a robust behavioral pattern but does not isolate its exact source.


\section*{Data Availability}
The EmpathicStories dataset is publicly available from \citet{shen2023empathicstories}. LLM-generated rewrites, analysis code, computed linguistic markers, and statistical results are available at \url{https://github.com/tomvannuenen/narrative-normalization}.

\section*{AI Disclosure Statement}
This manuscript was prepared with the assistance of Claude (Anthropic), used for Python code debugging and statistical verification. The author also used the tool to identify structural redundancies in the paper. Any model-generated text underwent substantial revision by the author. The empirical design, analysis, interpretation, and argumentation are entirely the author's own work.

\section*{Funding}
This research received no specific grant from any funding agency in the public, commercial, or not-for-profit sectors.

\section*{Conflict of Interest}
The author declares no conflict of interest.

\bibliographystyle{abbrvnat}
\bibliography{references}

\end{document}


\maketitle

\section{Core Marker Definitions}

The main paper reports findings for 13 markers selected for defensibility within computational stylistics and digital humanities. This section provides formal definitions for each marker.

\subsection{Function Words (2 markers)}

These markers follow the Delta tradition in authorship attribution, where function words are often treated as relatively topic-insensitive markers of stylistic patterning (Burrows, 2002; Eder et al., 2016).

\begin{description}
    \item[Function word ratio] Proportion of tokens belonging to closed-class grammatical categories (articles, prepositions, pronouns, auxiliaries, conjunctions) relative to total tokens. Computed using spaCy part-of-speech tags. Higher values indicate greater reliance on grammatical rather than lexical items.

    \item[MFW coverage] Proportion of tokens accounted for by the 50 most frequent words in the corpus. This metric captures reliance on very high-frequency vocabulary; higher values indicate greater concentration in common words. Based on the Most Frequent Words approach central to Burrows' Delta (Burrows, 2002).
\end{description}

\subsection{Vocabulary (5 markers)}

These markers capture lexical richness and character-level distributional patterning commonly used in stylometric analysis.

\begin{description}
    \item[Yule's K] A vocabulary concentration measure defined as:
    \[K = 10^4 \cdot \frac{\sum_{i=1}^{V} f_i^2 - N}{N^2}\]
    where $N$ is total tokens, $V$ is vocabulary size, and $f_i$ is the frequency of the $i$-th word type. Lower values indicate richer, less repetitive vocabulary (Yule, 1944). Relatively robust to text length.

    \item[Honor\'{e}'s R] A hapax-based richness measure defined as:
    \[R = 100 \cdot \frac{\log N}{1 - V_1/V}\]
    where $N$ is total tokens, $V$ is vocabulary size, and $V_1$ is the number of hapax legomena (words appearing exactly once). Higher values indicate greater vocabulary richness (Honor\'{e}, 1979).

    \item[MTLD] Measure of Textual Lexical Diversity. Computed as the mean length of sequential word strings maintaining a type-token ratio above 0.72. More robust to text length than traditional TTR (McCarthy \& Jarvis, 2010). Higher values indicate greater lexical diversity.

    \item[Mean word length] Mean number of characters per token. A commonly used stylometric measure (Mendenhall, 1887).

    \item[Character trigram entropy] Shannon entropy over character trigram distribution:
    \[H = -\sum_{t \in T} p(t) \log_2 p(t)\]
    where $T$ is the set of character trigrams and $p(t)$ is the probability of trigram $t$. Higher values indicate more diverse character-level patterns. Character n-grams can capture sublexical and orthographic patterning and are often less topic-sensitive than word-based features (Stamatatos, 2009).
\end{description}

\subsection{Syntax and Punctuation (3 markers)}

These markers capture clause structure and sentence-level patterns.

\begin{description}
    \item[Mean sentence length] Mean number of tokens per sentence. A standard complexity measure in stylometry (Williams, 1940; Stamatatos, 2009). Longer sentences often correlate with greater syntactic elaboration, though they may also reflect coordination and punctuation practices.

    \item[Comma frequency] Commas per 100 tokens. Captures clause segmentation, parenthetical insertions, and listing behavior. An established stylometric feature (Grieve, 2007).

    \item[Dash frequency] Dashes (em-dashes and en-dashes) per 100 tokens. Captures parenthetical and interruptive punctuation habits. Higher values indicate more frequent use of interruptive or parenthetical punctuation.
\end{description}

\subsection{Register (3 markers)}

These markers index formality along the written--spoken continuum (Biber, 1988; Heylighen \& Dewaele, 2002).

\begin{description}
    \item[Contraction density] Contractions (e.g., \textit{don't}, \textit{I'm}, \textit{can't}) per 100 tokens. A direct marker of informal register; reduction signals formalization. Based on Pennebaker \& King (1999).

    \item[First-person pronoun density] First-person pronouns (\textit{I}, \textit{me}, \textit{my}, \textit{mine}, \textit{we}, \textit{us}, \textit{our}, \textit{ours}) per 100 tokens. Higher values indicate more explicit first-person reference in the discourse. Central to studies of personal narrative (e.g. Biber, 1988; Pennebaker et al., 2003).
    
    \item[Emotion word density] Tokens matching NRC Emotion Lexicon entries (Mohammad \& Turney, 2013) per 100 tokens. Captures the density of explicitly affective lexical items.
\end{description}

\section{Statistical Notes}

\subsection{Effect Size Measures}

The analysis reports Cohen's $d$ as a standardized effect size for paired original--rewrite comparisons. As presented in tables and figures, positive $d$ indicates the feature increases in rewrites; negative $d$ indicates the feature decreases. This convention aligns the sign of the effect size with the direction of change.

\textbf{Rank-biserial correlation.} As a robustness check, rank-biserial correlation ($r_{rb}$) was also computed, a fully nonparametric effect-size measure appropriate for Wilcoxon signed-rank tests. The correlation between $d$ and $r_{rb}$ across all markers exceeded $r = 0.98$ for all models, confirming that the effect-size estimates are robust to distributional assumptions.

\subsection{Averaging}

Mean effect sizes reported in tables (e.g., Mean $|d|$) are \textbf{unweighted marker-level averages}. Each of the 13 markers contributes equally to the mean, regardless of which category it belongs to.

\subsection{Direction Agreement}

Direction agreement (Table 2 in main text) is computed as \textbf{within-model comparison}: for each model, each marker's direction under rewrite-only or voice-preserving is compared to that same model's direction under generic, then averaged across models. This measures whether prompt condition affects the direction of change for individual models, not cross-model consensus.

\section{Source-Level Analyses}

The EmpathicStories corpus draws from six sources with different baseline registers. To assess whether normalization patterns generalize across source types, stratified analyses were conducted on the three largest source categories.

\subsection{Baseline Register Differences}

\begin{table}[h]
\centering
\begin{tabular}{lcc}
\toprule
\textbf{Source} & \textbf{Contraction Rate} & \textbf{MTLD} \\
\midrule
Reddit & 0.042 & 48.2 \\
Hippocorpus & 0.028 & 52.1 \\
Oral History & 0.035 & 45.8 \\
\bottomrule
\end{tabular}
\caption{Supplementary Table S1: Baseline marker values by source category (means). On these two markers, Reddit narratives appear the most informal and Hippocorpus recollections the most formal at baseline.}
\label{tab:S1}
\end{table}

\subsection{Normalization by Source}

The core pattern observed in the main analysis holds across source types: lexical diversity increases and several first-person or informality-linked markers decrease. However, effect magnitudes differ:

\begin{itemize}
    \item \textbf{Reddit narratives} show the largest voice-marker deflation (mean $|d| = 0.52$ for voice dimension), consistent with their more informal baseline register.

    \item \textbf{Hippocorpus recollections} show smaller effects (mean $|d| = 0.28$ for voice dimension), consistent with a ceiling effect: texts already closer to formal written register have less room to shift.

    \item \textbf{Oral history transcripts} show intermediate effects, with contraction reduction comparable to Reddit (possibly because transcription conventions had already normalized some spoken features).
\end{itemize}

These differences suggest that the magnitude of change varies by source type rather than taking a uniform form across all inputs. One possible explanation is that the degree of change depends in part on the distance between the source register and the distribution of output favored by the model. The same rewriting operation that substantially transforms an informal Reddit post may produce only modest changes in a more formal recollection, even when the directional tendency is the same.

\subsection{Directional Consistency Across Sources}

To assess whether the pattern is better described as stylistic compression than as genre translation, effect sizes were computed separately within each source category. Table~\ref{tab:source_stratified} shows that 12 of 13 markers exhibit consistent directional shifts across all three major source types (Hippocorpus, Road Trip oral histories, Reddit posts).

\begin{table}[h]
\centering
\small
\begin{tabular}{llcccc}
\toprule
\textbf{Marker} & \textbf{Category} & \textbf{Hippo} & \textbf{Road Trip} & \textbf{Reddit} & \textbf{Consistent?} \\
\midrule
Comma frequency & Syntax & $+$ & $+$ & $+$ & Yes \\
MTLD & Vocabulary & $+$ & $+$ & $+$ & Yes \\
MFW coverage & Function Words & $-$ & $-$ & $-$ & Yes \\
Honor\'{e}'s R & Vocabulary & $+$ & $+$ & $+$ & Yes \\
Dash frequency & Syntax & $+$ & $+$ & $+$ & Yes \\
Trigram entropy & Vocabulary & $+$ & $+$ & $+$ & Yes \\
Mean word length & Vocabulary & $+$ & $+$ & $+$ & Yes \\
Function word ratio & Function Words & $-$ & $-$ & $-$ & Yes \\
Yule's K & Vocabulary & $-$ & $-$ & $-$ & Yes \\
First-person pronouns & Register & $-$ & $-$ & $-$ & Yes \\
Sentence length & Syntax & $+$ & $+$ & $+$ & Yes \\
Emotion words & Register & $+$ & $+$ & $+$ & Yes \\
\midrule
Contraction density & Register & $+$ & $-$ & $+$ & \textbf{No} \\
\bottomrule
\end{tabular}
\caption{Supplementary Table S2: Directional consistency of normalization effects across source categories (GPT-5.4, generic condition). $+$ = rewrite increases feature; $-$ = rewrite decreases feature. Twelve of thirteen markers show identical direction across all three source types, suggesting that the pattern is better characterized as compression of stylistic variation within source types than as simple translation between source types. The exception (contraction density) shows LLMs reducing contractions in high-contraction oral transcripts while slightly increasing them in low-contraction recollections. This pattern may suggest normalization toward a standard register from both directions.}
\label{tab:source_stratified}
\end{table}

\subsection{Word-Based Delta Robustness Check}

To confirm that the findings are not an artifact of the character n-gram feature space, the source--rewrite matching analysis was repeated using a word-based feature set more closely aligned with the Delta tradition, including MFW coverage, function word ratio, and vocabulary richness measures. Although Delta is generally considered better suited to longer texts (Eder, 2015), more recent work has also applied stylometric comparison to AI--human contrasts on texts of comparable length (O'Sullivan, 2025).

\begin{table}[h]
\centering
\begin{tabular}{lccc}
\toprule
\textbf{Model} & \textbf{Char n-grams} & \textbf{Word-based} & \textbf{Full features} \\
\midrule
GPT-5.4 & 14.3\% & 15.7\% & 17.7\% \\
Claude Sonnet & 1.7\% & 2.3\% & 0.7\% \\
Gemini 3.1 Pro & 1.0\% & 0.3\% & 0.7\% \\
\midrule
Chance baseline & 0.3\% & 0.3\% & 0.3\% \\
\bottomrule
\end{tabular}
\caption{Supplementary Table S3: Source-rewrite matching accuracy across feature spaces (generic prompt condition). Character n-gram features are often preferred for short texts; the word-based set provides a complementary comparison grounded in established stylometric practice; full features combines both sets. The pattern is consistent across all feature spaces: GPT-5.4 rewrites remain more recoverable to their source texts (above chance), while Claude and Gemini rewrites fall to near-chance levels.}
\label{tab:word_delta}
\end{table}

\section{Narrative Stance Markers}

In addition to the 13 stylometric markers reported in the main text, five markers were computed to capture two related dimensions of narrative stance: narrated positioning (where the narrator stands relative to events) and explanatory style (how understanding is conveyed). These markers draw on narrative theory (Labov \& Waletzky, 1967; Bruner, 1991) and discourse analysis (Biber \& Conrad, 2019).

\subsection{Marker Definitions}

\begin{description}
    \item[Eventive clause density] Proportion of clauses containing physical action, motion, or perception verbs (go, walk, run, see, hear, etc.). Higher values indicate more event-driven, situated narration.

    \item[First-person eventive density] First-person pronouns immediately followed by action verbs (e.g., ``I walked,'' ``I saw,'' ``we ran'') per sentence. Captures narrator-as-agent constructions often associated with experiential narration.

    \item[Causal connective density] Causal and explanatory connectives (because, therefore, since, as a result) per sentence. Indexes explicit causal reasoning in the narrative.

    \item[Retrospective framing density] Phrases explicitly marking temporal distance or hindsight (``looking back,'' ``in retrospect,'' ``I now realize,'' ``at the time I didn't know'') per sentence.

    \item[Abstraction density] Abstract nouns (concepts, states, qualities---typically words ending in -tion, -ness, -ity, etc.) per 100 words. Higher values indicate more nominalized and abstract prose.
\end{description}

Two additional markers---concrete noun density and sensory language density---were computed for exploratory purposes. Effect sizes for these markers were small ($d = -0.17$ and $d = -0.20$ respectively)---much smaller than the formalization effects reported in the main text, but contrary to what uniform flattening would predict. This suggests that rewrites may preserve some descriptive surface features even as experiential stance markers decline.

\subsection{Voice-Preserving Effects on Stance Markers}

Table~\ref{tab:stance_voice} compares generic and voice-preserving conditions for the five stance markers.

\begin{table}[h]
\centering
\small
\begin{tabular}{lccc}
\toprule
\textbf{Marker} & \textbf{Generic $d$} & \textbf{Voice $d$} & \textbf{Reduction} \\
\midrule
Eventive clause density & $-0.31$ & $-0.01$ & 97\% \\
Abstraction density & $+0.58$ & $+0.14$ & 76\% \\
First-person eventive & $-0.29$ & $-0.20$ & 31\% \\
Causal connectives & $-0.47$ & $-0.39$ & 17\% \\
Retrospective framing & $+0.49$ & $+0.33$ & 33\% \\
\bottomrule
\end{tabular}
\caption{Supplementary Table S4: Voice-preserving prompt effectiveness for stance markers (mean across models). Voice-preserving prompts substantially reduce effects on eventive clause density (97\%) and abstraction (76\%), but are less effective at preserving first-person eventive constructions, causal connectives, and retrospective framing (17--33\% reduction).}
\label{tab:stance_voice}
\end{table}

\subsection{Interpretation}

The stance marker analysis suggests that the observed shifts extend beyond surface register alone. Across these measures, rewrites show less eventive positioning, more retrospective framing, and greater abstraction. Taken together, these patterns are consistent with a movement away from situated narration and toward more compressed interpretation.

\section{Model-Specific Anomaly: Increased Compression Under Voice-Preserving Prompting}

While the main text reports that voice-preserving prompts reduce normalization magnitude by approximately 32\% on average, one model shows a counterintuitive pattern: for Claude Sonnet, voice-preserving instructions are associated with greater stylometric compression rather than reduced compression.

\subsection{Findings}

Using Levene's test for equality of variances across 58 stylometric features (combining the 13 core markers with additional character n-gram and lexical measures), variance compression was compared between prompt conditions:

\begin{itemize}
    \item \textbf{Generic prompt:} Claude compressed 78\% of features (45/58) with a median variance reduction of 26\%. Multivariate dispersion decreased by 9\% (Cohen's $d = 0.36$).

    \item \textbf{Voice-preserving prompt:} Claude compressed 95\% of features (55/58) with a median variance reduction of 34\%. Multivariate dispersion decreased by 16\% (Cohen's $d = 0.65$).
\end{itemize}

This represents an 80\% increase in effect size ($d = 0.36 \rightarrow 0.65$), moving from a ``small'' to a ``medium'' effect under the conventional benchmarks. The proportion of features showing compression rises to 95\%.

\subsection{Comparison Across Models}

GPT-5.4 and Gemini 3.1 Pro do not show this paradox. For both models, dispersion reduction remains stable or decreases slightly between generic and voice-preserving conditions:

\begin{table}[h]
\centering
\begin{tabular}{lccc}
\toprule
\textbf{Model} & \textbf{Generic $d$} & \textbf{Voice $d$} & \textbf{Change} \\
\midrule
GPT-5.4 & 0.22 & 0.22 & $-3\%$ \\
Claude Sonnet & 0.36 & 0.65 & $+80\%$ \\
Gemini 3.1 Pro & 0.22 & 0.25 & $+12\%$ \\
\bottomrule
\end{tabular}
\caption{Supplementary Table S5: Dispersion reduction effect sizes by model and prompt condition. Claude shows a substantial increase in compression under voice-preserving prompts; GPT remains stable and Gemini shows a modest increase.}
\label{tab:paradox}
\end{table}

\subsection{Interpretation}

One possible interpretation is that explicit instructions to preserve voice prime the model to attend to voice-related features while simultaneously normalizing them. The instruction may activate attention to stylistic dimensions that the model then ``corrects'' according to its implicit preferences. However, this mechanism is speculative; the data establish the behavioral pattern but do not isolate its cause.

The effect is model-specific, but its presence in one major system demonstrates that prompt-based mitigation strategies may not only fail to prevent normalization but could, in some cases, intensify it. This finding underscores the limits of user control over AI-mediated transformation.

\section{Perplexity Analysis}

If AI rewrites converge toward a common register, they may also become more predictable in distributional terms. Perplexity measures the average uncertainty a language model assigns to a text: lower perplexity indicates that the sequence is more readily predicted by the model, and therefore more typical under that model's learned distribution. Here, a decrease in perplexity would suggest that rewritten texts have shifted toward more model-typical language.

\subsection{Method}

Perplexity scores were computed using GPT-2 (124M parameters) as a reference model. For each text, the mean per-token log probability was computed and converted to perplexity. GPT-2 serves as an independent reference model distinct from the rewriting systems under study.

\subsection{Results}

All three AI systems produced rewrites with lower perplexity than the originals:

\begin{table}[h]
\centering
\begin{tabular}{lcccc}
\toprule
\textbf{Model} & \textbf{Original} & \textbf{Rewrite} & \textbf{\% Change} & \textbf{Cohen's $d$} \\
\midrule
GPT-5.4 & 22.8 & 17.7 & $-22.6\%$ & 1.18 \\
Claude Sonnet & 22.8 & 20.7 & $-9.4\%$ & 0.44 \\
Gemini 3.1 Pro & 22.8 & 21.8 & $-4.5\%$ & 0.19 \\
\bottomrule
\end{tabular}
\caption{Supplementary Table S6: Perplexity reduction under LLM rewriting (generic prompt condition). All models reduce perplexity; GPT-5.4 shows the largest effect.}
\label{tab:perplexity}
\end{table}

All effects were statistically significant ($p < 0.001$ for GPT-5.4 and Claude; $p = 0.001$ for Gemini).

\subsection{Interpretation}

The perplexity findings complement the stylometric analysis. Vocabulary diversity increases (as measured by MTLD and Honor\'{e}'s R), yet the distribution of that vocabulary becomes more predictable to a language model. Texts become simultaneously more elaborate and more generic, i.e., polished prose that sounds more like other polished prose. This pattern is consistent with the hypothesis that LLM rewriting draws texts toward a more standardized and model-typical register that is lexically varied but distributionally conventional.

\section*{References}

\small

Biber, D. (1988). \textit{Variation Across Speech and Writing}. Cambridge University Press.

Biber, D., \& Conrad, S. (2019). \textit{Register, Genre, and Style} (2nd ed.). Cambridge University Press.

Bruner, J. (1991). The narrative construction of reality. \textit{Critical Inquiry}, 18(1), 1--21.

Burrows, J. (2002). `Delta': A measure of stylistic difference and a guide to likely authorship. \textit{Literary and Linguistic Computing}, 17(3), 267--287.

Eder, M. (2015). Does size matter? Authorship attribution, small samples, big problem. \textit{Digital Scholarship in the Humanities}, 30(2), 167--182.

Eder, M., Rybicki, J., \& Kestemont, M. (2016). Stylometry with R: A package for computational text analysis. \textit{The R Journal}, 8(1), 107--121.

Grieve, J. (2007). Quantitative authorship attribution: An evaluation of techniques. \textit{Literary and Linguistic Computing}, 22(3), 251--270.

Heylighen, F., \& Dewaele, J.-M. (2002). Variation in the contextuality of language: An empirical measure. \textit{Foundations of Science}, 7(3), 293--340.

Honor\'{e}, A. (1979). Some simple measures of richness of vocabulary. \textit{Association for Literary and Linguistic Computing Bulletin}, 7(2), 172--177.

Labov, W., \& Waletzky, J. (1967). Narrative analysis: Oral versions of personal experience. In J. Helm (Ed.), \textit{Essays on the Verbal and Visual Arts} (pp. 12--44). University of Washington Press.

McCarthy, P. M., \& Jarvis, S. (2010). MTLD, vocd-D, and HD-D: A validation study of sophisticated approaches to lexical diversity assessment. \textit{Behavior Research Methods}, 42(2), 381--392.

Mendenhall, T. C. (1887). The characteristic curves of composition. \textit{Science}, 9(214), 237--249.

Mohammad, S. M., \& Turney, P. D. (2013). Crowdsourcing a word--emotion association lexicon. \textit{Computational Intelligence}, 29(3), 436--465.

O'Sullivan, J. (2025). Stylometric comparisons of human versus AI-generated creative writing. \textit{Humanities and Social Sciences Communications}, 12, 1708. https://doi.org/10.1038/s41599-025-05986-3

Pennebaker, J. W., Boyd, R. L., Jordan, K., \& Blackburn, K. (2015). The development and psychometric properties of LIWC2015. University of Texas at Austin.

Pennebaker, J. W., \& King, L. A. (1999). Linguistic styles: Language use as an individual difference. \textit{Journal of Personality and Social Psychology}, 77(6), 1296--1312.

Pennebaker, J. W., Mehl, M. R., \& Niederhoffer, K. G. (2003). Psychological aspects of natural language use: Our words, our selves. \textit{Annual Review of Psychology}, 54, 547--577.

Stamatatos, E. (2009). A survey of modern authorship attribution methods. \textit{Journal of the American Society for Information Science and Technology}, 60(3), 538--556.

Williams, C. B. (1940). A note on the statistical analysis of sentence-length as a criterion of literary style. \textit{Biometrika}, 31(3/4), 356--361.

Yule, G. U. (1944). \textit{The Statistical Study of Literary Vocabulary}. Cambridge University Press.